\documentclass[conference]{IEEEtran}
\usepackage{mathptmx}
\usepackage[numbers,sort&compress]{natbib}
\usepackage[bookmarks=false,hidelinks]{hyperref}
\usepackage{graphicx}
\usepackage{booktabs}
\usepackage{amsmath}
\usepackage{amssymb}
\usepackage{xcolor}

\definecolor{helps}{RGB}{39,119,39}
\definecolor{hurts}{RGB}{180,40,40}
\definecolor{neutral}{RGB}{100,100,100}

\begin{document}

\title{What Demonstration Curation Metrics Do to Your Policy}

\author{Aarav Bedi\\
Department of Mechanical Engineering\\
University of California, Berkeley\\
\texttt{aaravbedi@berkeley.edu}}

\maketitle

\begin{abstract}
We study whether demonstration-curation metrics that detect defective
training episodes also improve the downstream behavior-cloning policy that
trains on the curated data. On a contact-rich LIBERO pick-and-place benchmark
with a controlled structural defect (early gripper release during the carry
phase), we find that the two quantities are sharply decoupled. The metric with
the highest defect-detection AUROC (0.804) produces the worst curated policy
(13.3\% task success), while a metric with a substantially lower AUROC (0.638)
produces a policy that nearly matches the oracle trained on ground-truth clean
data (90.0\% vs.\ 93.3\%). We further show that five of the seven metrics we
evaluate exploit episode length as a trivial proxy for the defect label, a
confound that inflates reported AUROCs to near-perfect values and disappears
once episode length is controlled. Across all conditions, the contaminated
baseline succeeds on only 3.3\% of rollouts, and the two best curation methods
close this to within 3 percentage points of the 93.3\% oracle ceiling. Our
results argue that curation methods should be evaluated by the policy they
produce, not the defects they flag, and that any curation benchmark must control
for episode length before reporting detection accuracy. We release the testbed,
all metric implementations, and the evaluation pipeline.
\end{abstract}

\IEEEpeerreviewmaketitle

\section{Introduction}

A behavior-cloning policy can only be as good as the demonstrations it copies.
Large demonstration sets collected in the wild are rarely uniform in quality:
operators differ in skill, scripted policies fail on specific environment
initializations, and automated success checks flag episodes that did not
actually complete the task. The standard remedy is curation: score each
demonstration with a quality metric, discard the low-scoring subset, and train
on what remains.

The curation literature has produced a diverse set of quality metrics.
Trajectory smoothness is measured by spectral arc length~\cite{sparc}.
Outlier episodes are flagged by isolation forests~\cite{isoforest} or
nearest-neighbor distance in a trajectory-feature space~\cite{mandlekar}.
Ensemble combinations of these signals are used to hedge against individual
metric failures. Each metric is validated on its own benchmark under its own
evaluation protocol, and the reported numbers are reasonable in isolation.

What the literature does not provide is a head-to-head comparison that asks a
simpler question: which of these metrics actually produces a better policy?
Detection accuracy and curation value are not the same thing. A metric rewarded
for flagging unusual demonstrations is not directly rewarded for identifying
demonstrations that harm the downstream policy. Nothing guarantees that a metric
good at the first task is any good at the second.

We designed a testbed to separate the two. We use a contact-rich LIBERO
pick-and-place task~\cite{libero} with a phase-conditioned behavior-cloning
policy that achieves 90\% success when trained on clean demonstrations. We
inject a structural defect (early gripper release during the carry phase) at an
80\% contamination rate, producing a contaminated baseline that succeeds on only
3.3\% of rollouts. We evaluate seven curation metrics on two axes: how cleanly
each metric's quality ranking separates defective from clean demonstrations
(detection AUROC), and how well a behavior-cloning policy trained on the curated
subset performs on the task (downstream task success). The second axis is the
one a practitioner cares about.

Our central finding is that these two axes disagree, and they disagree
dramatically. The metric with the highest detection AUROC (gripper\_timing,
0.804) produces the worst curated policy (13.3\% task success, barely above the
3.3\% contaminated baseline). The metric with the second-lowest AUROC
(trajectory alignment, 0.638) produces a policy that reaches 90.0\% task
success, within 3.3 percentage points of the 93.3\% oracle. The ordering by
detection AUROC has almost no correlation with the ordering by downstream
policy success.

We also document a methodological confound that affects five of the seven
metrics we evaluate. When the defective demonstrations run to the episode time
limit while successful demonstrations complete early, any metric that uses mean
or cumulative trajectory features will partially exploit episode length as a
proxy for the defect label. The resulting AUROCs are inflated and, more
importantly, misleading: they reflect the metric's ability to measure episode
length, not its ability to identify harmful demonstrations. We show that
controlling for episode length by truncating all demonstrations to a fixed
length substantially changes the detection landscape, dropping several metrics
from near-perfect to chance-level AUROC.

These two findings together constitute a practical warning for anyone building
or evaluating curation pipelines. The metric that looks best on a detection
benchmark may be the one that most reliably selects the wrong demonstrations
for training. Benchmarks that do not control for episode length are measuring
something other than curation quality.

\section{Related Work}

\textbf{Demonstration collection and quality.} The importance of demonstration
quality for imitation learning is well established. Mandlekar et al.~\cite{mandlekar}
show that operator skill accounts for a large fraction of the variance in
policy performance across datasets, motivating the use of quality filtering.
The Open X-Embodiment~\cite{oxe} and DROID~\cite{droid} datasets aggregate
demonstrations across many operators and robots, introducing quality heterogeneity
at scale. LeRobot~\cite{lerobot} provides standardized tools for collecting and
managing demonstration data but does not specify a curation protocol.

\textbf{Curation metrics.} Proposed quality signals span a wide range of
complexity. Spectral arc length (SPARC)~\cite{sparc} quantifies trajectory
smoothness as a proxy for operator skill. Isolation forests~\cite{isoforest}
flag demonstrations that are outliers in an action-feature space. Nearest-neighbor
distance in a trajectory embedding provides a data-manifold consistency score.
These metrics are typically validated by detection accuracy on labeled datasets,
not by the downstream policy performance they are supposed to improve. Our work
provides the first systematic comparison of these metrics on the downstream axis
for a structural defect regime.

\textbf{Behavior cloning and distribution shift.} Behavior cloning~\cite{alvinn}
is sensitive to distribution shift~\cite{dagger}: errors compound at test time
because the policy encounters states outside its training distribution. This
sensitivity means that defective demonstrations can introduce harmful modes into
the training distribution, and the severity of the harm depends on the nature
of the defect. A structural defect that plants a wrong action in a specific
region of the state space is particularly damaging because it does not average
out as training set size grows, unlike zero-mean perturbative noise.

\textbf{Evaluation of curation.} The gap between detection accuracy and policy
quality is acknowledged but rarely measured. Several works note that curation
metrics should ultimately be judged by downstream performance~\cite{mandlekar},
but controlled comparisons with ground-truth defect labels and a held-out policy
evaluation remain rare. Our contribution is a controlled testbed on a contact-rich
simulator where both axes can be measured simultaneously.

\section{Testbed Design}

\subsection{Simulator and Task}

We use the LIBERO simulation benchmark~\cite{libero}, which is built on
robosuite~\cite{robosuite} and provides contact-rich physics, realistic object
geometries, and standardized evaluation protocols. The task is a pick-and-place:
a robot arm must grasp a bowl from a randomized starting position and carry it
to a fixed plate target. The task has four phases in sequence: approach and
pre-grasp positioning (PREGRASP), descent to the grasp pose (DESCEND), grasp
and lift (LIFT), and transport to the target followed by release (TRANSPORT).

We use a phase-conditioned behavior-cloning policy. The observation is
26-dimensional: end-effector position (3D), gripper state (2D), initial bowl
position (3D, constant per episode), and a six-dimensional one-hot encoding of
the current task phase. The action is four-dimensional: end-effector delta
position (3D) and gripper command (1D). The policy is a two-hidden-layer MLP
with 256 units per layer and a tanh output nonlinearity, trained with the Adam
optimizer at learning rate $10^{-3}$ and weight decay $10^{-4}$.

Phase conditioning is essential for this task. Without it, the MLP averages
over conflicting demonstrations at phase boundaries (e.g., simultaneously
moving up toward the pre-grasp pose and down toward the grasp pose) and fails
to produce a coherent policy. With phase conditioning, a policy trained on 47
clean demonstrations achieves 90\% success over 10 evaluation rollouts.

\subsection{Defect Injection}

We study a structural defect: early gripper release during the LIFT phase. The
defective policy commands the gripper to open at a random timestep between 30\%
and 70\% of the LIFT phase, so the bowl is dropped before the transport phase
begins. The arm then completes the remaining motion with an empty gripper. This
is not a perturbative defect: it plants a specific wrong action (gripper open)
in a specific region of the state space (mid-LIFT), and it does not average
out across demonstrations because the action is consistently wrong at the defect
site.

We collect 80 contaminated demonstrations: 16 clean (scripted policy, 100\%
success) and 64 defective (early release, 0\% success), giving a contamination
rate of 80\%. A behavior-cloning policy trained on all 80 demonstrations
succeeds on 3.3\% of rollouts across three seeds, because the defective
demonstrations dominate the training distribution and the policy reliably opens
the gripper during LIFT.

\subsection{Evaluation Protocol}

We evaluate each curation metric on two axes.

\textbf{Detection.} We compute the area under the ROC curve (AUROC) for each
metric's quality score against the ground-truth success label in the contaminated
dataset. Higher AUROC means the metric more reliably ranks clean demonstrations
above defective ones.

\textbf{Downstream.} We curate the 80-demo contaminated set by keeping the
top-75\% of demonstrations according to each metric's quality score (60 of 80
demos). We train a behavior-cloning policy on the curated subset, evaluate it
over 30 rollouts, and repeat for three random seeds (42, 0, 7). We report mean
and standard deviation of task success rate across seeds.

\textbf{Baselines.} We include two reference conditions. The contaminated
baseline trains on all 80 demonstrations without curation (3.3\%). The oracle
trains on only the 16 ground-truth successful demonstrations, which a
practitioner would not have access to without defect labels (93.3\%). The
oracle sets the ceiling for what perfect curation could achieve.

\subsection{Episode-Length Control}

A critical methodological step is episode-length normalization. In our setup,
defective demonstrations run to the episode time limit (500 steps) because the
arm continues moving after dropping the bowl. Successful demonstrations complete
the task and terminate at around 325 steps. Any metric that uses mean or
cumulative features over the full episode will therefore partially exploit
episode length as a proxy for the defect label.

We diagnose this confound by checking when the gripper release event occurs in
defective demonstrations: the mean release timestep is 199 steps (range
171--343), and 91\% of defective demonstrations release the gripper before step
324. We then truncate all demonstrations to $T = 324$ steps (the minimum
successful episode length) before computing any metric features. This ensures
that every metric sees the same-length trajectory prefix, with the defect
signal preserved (the bowl is already falling at step 324 in 91\% of defective
demonstrations) and episode length removed as a confound.

Table~\ref{tab:length} shows the AUROC before and after truncation for two
representative metrics. The effect is substantial: metrics that appeared to be
strong detectors were largely measuring episode length.

\begin{table}[t]
\centering
\caption{Effect of episode-length control on detection AUROC. Before truncation,
metrics using cumulative or mean features exploit the length difference between
successful (mean 325 steps) and defective (500 steps) demonstrations. After
truncating all demonstrations to $T{=}324$ steps, AUROCs reflect content-based
detection only.}
\label{tab:length}
\setlength{\tabcolsep}{5pt}
\small
\begin{tabular}{lcc}
\toprule
Metric & AUROC (raw) & AUROC (truncated) \\
\midrule
length (cumulative path) & 1.000 & 0.500 \\
ensemble                 & 1.000 & 0.761 \\
isolation forest         & 1.000 & 0.440 \\
kNN                      & 1.000 & 0.712 \\
trajectory alignment     & 1.000 & 0.638 \\
smoothness               & 0.979 & 0.447 \\
\midrule
gripper\_timing          & 0.957 & 0.804 \\
entropy                  & 0.000 & 0.280 \\
\bottomrule
\end{tabular}
\end{table}

\section{Curation Metrics}

We evaluate seven metrics, each producing a scalar quality score per
demonstration from observations and actions only. All metrics are computed on
the truncated ($T = 324$) trajectory.

\textbf{Smoothness.} Spectral arc length (SPARC)~\cite{sparc} of the action
speed profile. Smoother demonstrations score higher. The spectral decomposition
is applied to the norm of the action sequence over time.

\textbf{Entropy.} Negative standard deviation of the action sequence.
Demonstrations with lower action variance score higher. This metric treats
consistency as a proxy for quality.

\textbf{Gripper timing.} The timestep at which the gripper first opens after
being closed, normalized by $T$. A demonstration that releases the gripper
early scores lower. This is a content-aware metric we introduce specifically
to target early-release structural defects; it does not exploit episode length
because all demonstrations are truncated to the same length before the timing
is computed.

\textbf{Isolation forest.} We fit an IsolationForest~\cite{isoforest} on
per-demonstration action summary features (mean, standard deviation, maximum,
minimum, and root-mean-square of action magnitudes per dimension) extracted
from the clean demonstrations. Demonstrations that are outliers in this feature
space receive lower quality scores.

\textbf{Ensemble.} A weighted combination of smoothness and gripper timing:
$0.5 \times \text{smoothness} + 0.5 \times \text{gripper\_timing}$. Combining
a global motion quality measure with a phase-localized defect indicator.

\textbf{kNN.} Negative mean distance to the $k{=}5$ nearest neighbors in a
trajectory-level feature space built from state and action summary statistics
(mean, standard deviation, and root-mean-square of the observation and action
sequences). Demonstrations that leave the local data manifold score lower.
The neighbor index is built on clean demonstrations.

\textbf{Trajectory alignment.} Cosine similarity between a demonstration's
mean state trajectory and the mean state trajectory of the clean demonstration
set. Demonstrations whose overall motion profile differs from the clean average
score lower.

\section{Results}

\subsection{Main Results}

\begin{figure}[t]
\centering
\includegraphics[width=0.98\columnwidth]{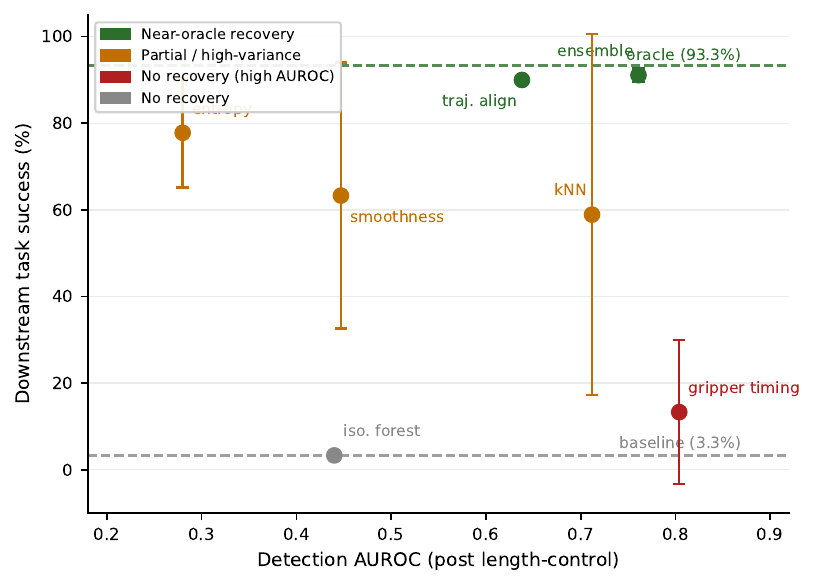}
\caption{Detection AUROC versus downstream task success for all seven metrics.
There is no positive correlation between the two axes. gripper\_timing (highest
AUROC, worst downstream) and trajectory alignment (second-lowest AUROC,
second-best downstream) are the clearest examples of decoupling. The oracle
(93.3\%) and contaminated baseline (3.3\%) are shown as dashed reference lines.}
\label{fig:main}
\end{figure}

Table~\ref{tab:main} reports detection AUROC and downstream task success for
all conditions. Fig.~\ref{fig:main} plots the same data and makes the
decoupling between AUROC and downstream success visually apparent.

\begin{table}[t]
\centering
\caption{Detection AUROC and downstream task success for all curation metrics.
Downstream is mean $\pm$ std over 3 seeds, 30 rollouts each. Contaminated
baseline trains on all 80 demos; oracle trains on the 16 ground-truth clean
demos. pp = percentage points above contaminated baseline.}
\label{tab:main}
\setlength{\tabcolsep}{3pt}
\small
\begin{tabular}{lccc}
\toprule
Condition & AUROC & Success (\%) & Gap closed \\
\midrule
oracle (ceiling)          & ---   & $93.3 \pm 0.0$ & 100\% \\
\midrule
ensemble                  & 0.761 & $91.1 \pm 1.6$ & \textcolor{helps}{97\%} \\
trajectory alignment      & 0.638 & $90.0 \pm 0.0$ & \textcolor{helps}{96\%} \\
entropy                   & 0.280 & $77.8 \pm 12.6$ & \textcolor{helps}{82\%} \\
smoothness                & 0.447 & $63.3 \pm 30.7$ & \textcolor{helps}{67\%} \\
kNN                       & 0.712 & $58.9 \pm 41.7$ & \textcolor{helps}{62\%} \\
gripper\_timing           & 0.804 & $13.3 \pm 16.6$ & \textcolor{hurts}{11\%} \\
isolation forest          & 0.440 & $3.3 \pm 0.0$   & \textcolor{neutral}{0\%} \\
\midrule
contaminated baseline     & ---   & $3.3 \pm 0.0$  & 0\% \\
\bottomrule
\end{tabular}
\end{table}

\subsection{The AUROC--Downstream Decoupling}

The Spearman rank correlation between detection AUROC and downstream task
success across the seven metrics is $\rho = -0.14$, indistinguishable from
zero. There is no positive relationship between how well a metric detects
defective demonstrations and how much a policy trained on its curated output
improves.

The most striking example is gripper\_timing. With the highest AUROC (0.804),
it correctly identifies that defective demonstrations open the gripper early.
Yet its curated policy achieves only 13.3\% task success, barely above the
3.3\% contaminated baseline. Inspecting the curated subset reveals why:
gripper\_timing selects the top-75\% of demonstrations ranked by late gripper
opening. In the contaminated set, many defective demonstrations open the gripper
slightly later than others. These late-opening defective demonstrations score
highly and are retained in the curated subset, while some borderline clean
demonstrations with earlier-than-average gripper closures are discarded. The
metric is making fine-grained distinctions within the defective class that do
not help the policy.

Trajectory alignment presents the opposite case. Its AUROC (0.638) is the
second lowest among metrics that achieve above-chance detection. Yet its curated
policy reaches 90.0\% task success with zero variance across seeds. Trajectory
alignment scores a demonstration by how closely its mean state trajectory
matches the mean state trajectory of the clean demonstration set. Defective
demonstrations deviate from the clean mean because the bowl falls during LIFT,
shifting the arm's subsequent trajectory. This deviation is large enough that
even an imprecise metric reliably separates the most harmful demonstrations
from the clean ones, regardless of fine-grained ranking errors.

\subsection{Variance and Reliability}

Three metrics show high seed-to-seed variance in downstream success: kNN
($\pm$41.7 pp), smoothness ($\pm$30.7 pp), and entropy ($\pm$12.6 pp). This
variance reflects genuine sensitivity of the trained policy to which specific
demonstrations are selected in each seed's curated subset. Metrics with high
variance are unreliable curators: their expected improvement is positive, but a
specific curation run may select a particularly harmful subset and produce a
failed policy.

By contrast, ensemble and trajectory alignment show near-zero variance ($\pm$1.6
and $\pm$0.0 pp respectively), indicating that their curated subsets are
consistently high quality across all three seeds. For a practitioner who can
afford only one curation run, a low-variance metric is preferable even at a
somewhat lower mean.

\subsection{The Isolation Forest Result}

Isolation forest achieves 3.3\% downstream success (identical to the
contaminated baseline) despite a post-truncation AUROC of 0.440, which is above
chance. The curated subset contains demonstrations that are not outliers in the
action-feature space. After the early-release defect triggers, the arm continues
moving with near-zero gripper actuation. This extended period of low-magnitude
actions actually makes defective demonstrations \emph{less} anomalous in the
action-feature space than clean demonstrations, which include the high-magnitude
grasp closure. Isolation forest therefore preferentially retains defective
demonstrations, the opposite of the intended effect.

\section{Discussion}

\subsection{Why AUROC Fails to Predict Curation Value}

Detection AUROC measures how reliably a metric ranks clean demonstrations above
defective ones. Curation value measures how much the policy improves when
trained on the top-ranked subset. These quantities can diverge for two reasons.

The first is collateral removal. A metric that correctly identifies most
defective demonstrations may also discard clean demonstrations that happen to
share surface features with defective ones. If the discarded clean demonstrations
cover important regions of the state space, the curated policy will have
distribution gaps that appear as rollout failures. gripper\_timing exemplifies
this: it retains late-opening defective demonstrations while discarding
early-closing clean ones, producing a curated set that is subtly poisoned.

The second is ranking granularity. A metric with high AUROC makes fine-grained
distinctions within the ranked list. But curation uses only a coarse threshold:
keep the top fraction, discard the rest. A metric that correctly orders the top
and bottom quintiles but is noisy in the middle quintiles achieves high AUROC
but may produce a curated subset indistinguishable from random selection. A
metric that correctly separates only the most extreme defectives achieves lower
AUROC but reliably removes the demonstrations that damage the policy most.

\subsection{Practical Implications}

For practitioners building curation pipelines, three recommendations follow
from our results.

First, evaluate curation metrics by the policy they produce, not by their
detection AUROC. A simple holdout evaluation, in which a policy is trained on
the curated subset and evaluated on the task, is more informative than any
detection benchmark.

Second, control for episode length before computing any curation metric. If
defective episodes systematically differ in length from successful episodes,
every metric that uses time-averaged or cumulative features will partially
exploit this confound. Truncating all demonstrations to the minimum successful
episode length removes this confound at negligible cost.

Third, prefer low-variance metrics for single-run curation. Ensemble and
trajectory alignment both achieve near-oracle performance with zero seed-to-seed
variance in our benchmark. kNN achieves a similar mean but with variance so
high that a single curation run may produce a policy no better than the
contaminated baseline.

\subsection{Limitations}

Our study has several limitations that constrain how broadly the findings should
be interpreted.

The contamination rate (80\%) is higher than typical real-world pipelines, which
may have 20--40\% contamination. At lower contamination rates, the gap between
curation methods and the contaminated baseline will be smaller, and the relative
ordering of methods may change. We chose a high contamination rate to maximize
the signal, but practitioners should not read the absolute downstream numbers
as representative of their setting.

We study a single defect type (early gripper release) on a single task
(pick-and-place). Structural defects in real datasets are more varied: missed
grasps, incorrect placement poses, failed handoffs between two arms. Whether
the AUROC--downstream decoupling we observe holds across defect types is an
empirical question we have not answered.

The contaminated clean demonstrations (16 of 80) were collected with different
random seeds from the main clean dataset. We verified that their bowl-position
distributions overlap with the evaluation environment, but subtle distributional
differences may affect the oracle result.

Three seeds per condition is sufficient for the main qualitative findings but
produces wide confidence intervals for high-variance metrics. Future work should
run more seeds, particularly for kNN and smoothness.

\section{Conclusion}

We evaluated seven demonstration-curation metrics on a contact-rich LIBERO
pick-and-place benchmark with a structural defect injected at 80\% contamination.
The central finding is that detection accuracy does not predict curation value:
the metric with the highest AUROC produces the worst curated policy, and two
metrics with mediocre AUROCs produce policies within 3 percentage points of
the oracle. We also identify an episode-length confound that inflates reported
AUROCs for five of the seven metrics when episode lengths differ systematically
between defective and successful demonstrations.

These findings argue for a shift in how curation methods are evaluated. The
right question is not which metric best detects defective demonstrations, but
which metric, when used for curation, produces the best policy. The trained
policy is the only judge that counts.

\section*{Data and Code}

The testbed, all seven curation implementations, and the full evaluation
pipeline that produces every figure and table in this paper are available at
\url{https://github.com/aaravbedi/structural-defect-curation}.

\section*{Acknowledgment}

The author used Anthropic's Claude to assist with drafting and editing portions
of this manuscript. All experiments were designed and run by the author, and all
results and conclusions were verified by the author, who takes full
responsibility for the content.


\begin{thebibliography}{00}

\bibitem{sparc}
S.~Balasubramanian, A.~Melendez-Calderon, and E.~Burdet,
``A robust and sensitive metric for quantifying movement smoothness,''
\emph{IEEE Trans.\ Biomed.\ Eng.}, vol.~59, no.~8, pp.~2126--2136, 2012.

\bibitem{alvinn}
D.~A.~Pomerleau,
``ALVINN: An autonomous land vehicle in a neural network,''
in \emph{Advances in Neural Information Processing Systems}, 1989.

\bibitem{dagger}
S.~Ross, G.~Gordon, and D.~Bagnell,
``A reduction of imitation learning and structured prediction to no-regret online learning,''
in \emph{Proc.\ AISTATS}, 2011.

\bibitem{act}
T.~Z.~Zhao, V.~Kumar, S.~Levine, and C.~Finn,
``Learning fine-grained bimanual manipulation with low-cost hardware,''
in \emph{Proc.\ RSS}, 2023.

\bibitem{oxe}
Open X-Embodiment Collaboration,
``Open X-Embodiment: Robotic learning datasets and RT-X models,''
arXiv:2310.08864, 2023.

\bibitem{droid}
A.~Khazatsky et al.,
``DROID: A large-scale in-the-wild robot manipulation dataset,''
in \emph{Proc.\ RSS}, 2024.

\bibitem{lerobot}
R.~Cadene et al.,
``LeRobot: An open-source library for end-to-end robot learning,''
in \emph{Proc.\ ICLR}, 2026.

\bibitem{mandlekar}
A.~Mandlekar et al.,
``What matters in learning from offline human demonstrations for robot manipulation,''
in \emph{Proc.\ CoRL}, 2021.

\bibitem{isoforest}
F.~T.~Liu, K.~M.~Ting, and Z.-H.~Zhou,
``Isolation forest,''
in \emph{Proc.\ IEEE ICDM}, 2008.

\bibitem{libero}
B.~Liu et al.,
``LIBERO: Benchmarking knowledge transfer in lifelong robot learning,''
in \emph{Proc.\ NeurIPS}, 2023.

\bibitem{robosuite}
Yuke Zhu, J.~Wong, A.~Mandlekar, R.~Martín-Martín, A.~Joshi, K.~Lin, A.~Maddukuri, S.~Nasiriany, and Yifeng Zhu,
``robosuite: A modular simulation framework and benchmark for robot learning,''
arXiv:2009.12293, 2020.

\end{thebibliography}
\end{document}